\documentclass{article}

\usepackage[frozencache,cachedir=minted-cache]{minted} 

\usepackage[preprint]{neurips}

\usepackage[utf8]{inputenc} 
\usepackage[T1]{fontenc}    
\usepackage{hyperref}       
\usepackage{url}            
\usepackage{booktabs}       
\usepackage{amsfonts}       
\usepackage{nicefrac}       
\usepackage{microtype}      
\usepackage{xcolor}         
\hypersetup{
    colorlinks,
    linkcolor={blue!50!black},
    citecolor={blue!50!black},
    urlcolor={blue!50!black},
}

\usepackage{graphicx}
\usepackage{placeins}
\usepackage{subcaption}
\usepackage{amsmath}
\usepackage{amssymb}
\usepackage{float}
\usepackage{dcolumn}
\usepackage{algorithmic}
\usepackage{algorithm}
\usepackage{natbib}
\usepackage{balance}
\usepackage{multirow}
\usepackage{multirow}
\usepackage{dcolumn}
\usepackage{multicol}
\newcolumntype{d}{D{.}{.}{-1}}
\usepackage{textcomp}
\usepackage{cleveref}
\usepackage{wrapfig}
\usepackage[nottoc]{tocbibind}
\usepackage{todonotes}
\usepackage{xspace}
\usepackage{fdsymbol}

\def\R{\mathbb{R}}
\def\C{\mathbb{C}}

\def\lasso{GSS\xspace}

\newcommand{\gelu}{$\mathrm{GELU}$ }
\renewcommand{\bar}{\overline}    
\newcommand{\elemexp}{\mathrm{elementwise}\text{-}\mathrm{exp}}
\newcommand{\dss}{\textsc{DSS} }
\newcommand{\dssexp}{\textsc{DSS}\textsubscript{\textsc{exp}} }

\renewcommand{\epsilon}{\varepsilon}

\newcommand\comment[1]{}                
\newcommand{\harshnote}[1]{\todo[color=red!10]{Harsh: #1}}
\newcommand\ag[1]{\textcolor{blue}{[AG: #1]}}

\newcommand{\ackharshsection}{\section*{Acknowledgments}}
\NewEnviron{ack_harsh}{%
  \ackharshsection
  \BODY
}
\DeclareTextFontCommand\texttt{\fontfamily{cmss}\selectfont}
\DeclareTextFontCommand\textss{\fontfamily{cmtt}\selectfont}

\title{Long Range Language Modeling via \\ Gated State Spaces}

\author{%
Harsh Mehta$^1$\thanks{
\ $^1$Google Research,\ \ $^2$Tel Aviv University,\ \ $^3$Boston University. 
\newline\phantom{x}\hspace{3ex}
{\texttt {\scriptsize harshm@google.com,\;ankitgupta.iitkanpur@gmail.com,\;ashok@cutkosky.com,\;neyshabur@google.com}}.
}  \And Ankit Gupta$^2$ \And Ashok Cutkosky$^3$ \And Behnam Neyshabur$^1$

}

\begin{document}

\maketitle

\begin{abstract}
\comment{State space models have shown to be effective at modeling long range dependencies, specifically on sequence classification tasks. In this work we focus on autoregressive sequence modeling over natural language, Github source code and ArXiv mathematics articles. Based on a few recent developments around the effectiveness of gated activation functions, we propose a new layer named \textit{Gated State Space} (\lasso) and show that it trains significantly faster than the diagonal version of S4 (i.e. DSS) on TPUs, is simple to implement and fairly competitive with several well-tuned Transformer-based baselines. Finally, we show that interleaving traditional Transformer blocks with \lasso improves performance even further.}

State space models have shown to be effective at modeling long range dependencies, specially on sequence classification tasks. In this work we focus on autoregressive sequence modeling over English books, Github source code and ArXiv mathematics articles. Based on recent developments around the effectiveness of gated activation functions, we propose a new layer named \textit{Gated State Space} (\lasso) and show that it trains significantly faster than the diagonal version of S4 (i.e. DSS) on TPUs, is fairly competitive with several well-tuned Transformer-based baselines and exhibits zero-shot generalization to longer inputs while being straightforward to implement. Finally, we show that leveraging self-attention to model local dependencies improves the performance of \lasso even further.
\end{abstract}


\section{Introduction}

Modeling long range dependencies on sequential data is a crucial step towards closing the gap with human-level performance on many tasks. Attention based models like Transformer \citep{vaswani2017attention} have proven to be a strong choice of backbone architecture for a considerable number of tasks across modalities and scale \citep{devlin2018bert,brown2020language,dosovitskiy2021an_vit}. Vanilla Multi-Head-Attention famously incurs $\Omega(L^2)$ penalty in modeling a sequence of length $L$. This is prohibitive at best for tasks where the model is required to capture long range dependencies from various parts of the input. Over the years, a variety of improvements have been proposed to alleviate this quadratic complexity (cf. \citep{tay2020efficient}). 

On a somewhat orthogonal direction, attention-free models based on state spaces, such as S4 \citep{gu2022efficiently} and DSS \citep{dss_gupta2022diagonal}, have shown remarkable improvements on Long Range Arena (LRA) \citep{tay2020long}, a benchmark designed with long range modeling as its focus and consists of diverse tasks with 1k-16k sequence length across modalities. These models require careful initialization, originally borrowing ideas from the theory of HiPPO matrices \citep{Voelker2019LegendreMU,gu2020hippo}, to achieve good results on LRA.

In this work, we explore and extend the use of state space models by focusing solely on the task of autoregressive sequence modeling \citep{brown2020language,gopher_rae2021scaling,chowdhery2022palm,zhang2022opt175b,chinchilla_hoffmann2022training,bigbench_srivastava2022imitation}. Several key properties endowed by the state space model family makes it particularly attractive, to at least fully explore it, in the context of language modeling. First, it reduces the $\Omega(L^2)$ complexity on input sequence length to $O(L\log L)$. This complexity results from the use of Fast Fourier Transform (FFT) \citep{Cooley1965AnAF} for performing convolutions. We will describe this in detail in later sections. Second, the state space model is fully parallelizable in the length dimension. This is an arguably subtle but an important property at training time. Note that transformers are also fully parallelizable, a worthy advantage over traditional RNNs for modeling sequences, which otherwise incurs only an $O(L)$ penalty. While this parallelism is useful at training time, it may also be a curse at inference time where decoding every token requires attending to the whole past. The ideal model is parallelizable at training time but incurs a small constant cost (per decoded token) at inference time. This brings us to the final point. Due to the inherent convolution-recurrence equivalence of the state space model, it can be made to accumulate state and unroll like an RNN at inference time without any approximations. 

\begin{figure*}[t]
\begin{minipage}{0.4\linewidth}
    \includegraphics[width=\linewidth]{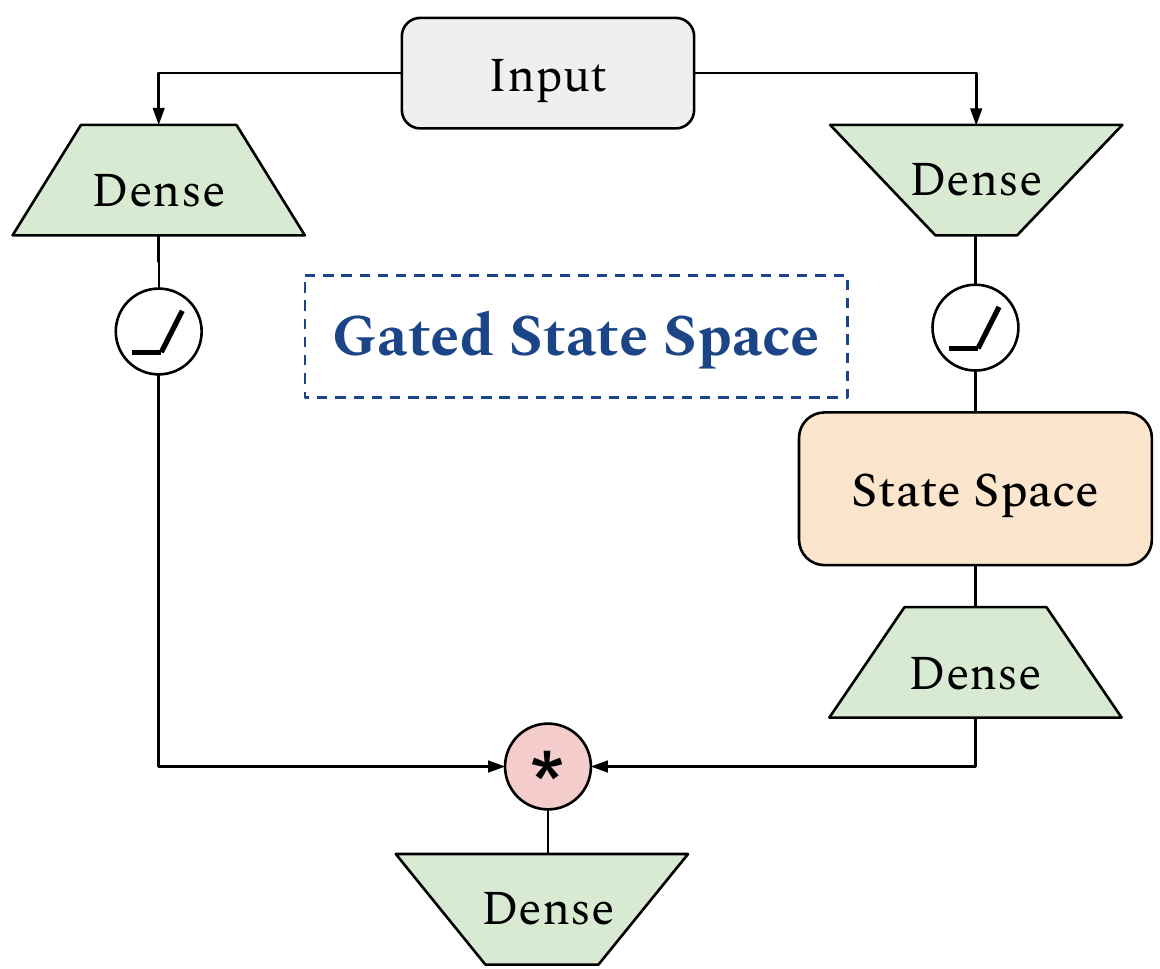}
\end{minipage}
\hfill
\qquad
\begin{minipage}{0.6\linewidth}
\begin{minted}
[
% frame=lines,
framesep=2mm,
baselinestretch=1.1,
fontsize=\footnotesize,
% linenos
]
{python}
    
def gss(x, F=4096, L=4096, E=1024, H=256):
    shortcut, x = x, norm(x)
    v = dense(x, F, activation='gelu')
    u = dense(x, H, activation='gelu') 
    y = dss(u, H, L)  
    # yh1,..,yhL are linear in uh1,..,uhL
    uc = dense(y, F)
    o = dense(uc * v, E)
    return o + shortcut

\end{minted}
\end{minipage}\setlength{\belowcaptionskip}{-13pt}
\caption{(a) Our proposed Gated State Space (\lasso) layer, (b) Pseudocode for \lasso (full implementation in \S\ref{app:code}).}\label{fig:lasso}
\end{figure*}

Despite these attractive properties, we found that current state space models (such as S4, DSS) run slower than we expected at training time on TPUs, our accelerator of choice. We take this opportunity to modify the architecture to reduce dimensionality of specific operations which we found to be bottlenecks. Our proposed changes borrow from a well-supported empirical observation around the effectiveness of gating units \citep{Shazeer2020GLUVI}. Specifically, \cite{flash_hua2022transformer} observed that replacing the typical Feed-Forward layer in the Transformer with gating units allows for a reduced dimensionality when mixing tokens along the length dimension using self-attention. We extend the use of gating units to state space model family and observe that, even in our context, the use of gating units allows for a reduction in dimensionality when performing FFT operations, which we observed to be the main bottleneck behind slow training. Furthermore, somewhat contrary to observations made by S4 and DSS authors, we found the performance of the model on language modeling tasks to be much less sensitive to initialization. We found that only the scale and structural aspects of initialization of state space variables were important and not the exact values. We were able to successfully train the model while initializing the state space variables randomly. This departs significantly, at least in understanding, from the reliance of the design on the theory of HiPPO matrices, which led the S4 model to employ several numerical linear algebra tricks to able to make it work. Combining both of these contributions, we propose a layer named Gated State Space (\lasso) (Figure \ref{fig:lasso}), which we empirically verified to be 2-3$\times$ faster than DSS while keeping the perplexity on several language modeling benchmarks (Table \ref{tab:dss_lasso}).

Going one step further, we also perform an apples-to-apples comparison with well-tuned and performant baselines reported in Block Recurrent Transformers \citep{hutchins2022block}, on several long range language modeling benchmarks over modalities such as English books, raw source code from Github and LaTeX source of ArXiv mathematics articles. As detailed in Table \ref{tab:other}, while our \lasso model currently lags behind on some tasks when compared in the fixed-parameter setting, it is fairly competitive in the \textit{fixed-compute} setting where we measure compute as the exact amount of TPUv4 hours spent on a training run and serves as a fairly accurate proxy to the realistic cost of training that model. Furthermore, we also experimented with a hybrid model in which we sparingly interleave Transformer layers (having local attention) in a \lasso stack to allow for a richer modeling of short range interactions. To our delight, this further improves performance at (roughly) no extra training cost, both in terms of parameters and compute.

While in our experiments we train on sequences of length at most 4k, we evaluated our \lasso variants on a wide range of sequence lengths upto 65k and found consistent generalization to longer inputs. Not only the performance doesn't degrade as the sequence length is increased but it gets significantly better, suggesting that \lasso is effective at utilizing the extra context even though it was not trained with that much amount of context.

At inference time, state space models including \lasso are fairly efficient since decoding can happen in recurrent mode (as much as $60\times$ better in the case of S4 \citep{gu2022efficiently}). Though, the hybrid model which also uses local attention complicates this advantage a bit. 

In summary, we propose \lasso, an alternative to S4 and \dss which trains 2-3$\times$ faster, is simple to implement and fairly competitive with well-tuned Transformer-based baselines on several long range language modeling benchmarks.













\section{Related Work}

In recent years, attention-based models have emerged as a dominant technique for sequence modeling, achieving remarkable improvements in a wide range of tasks, starting in NLP \citep{vaswani2017attention, devlin2018bert, radford2019language, liu2019roberta}, then moving to other classical machine learning areas such as computer vision \citep{dosovitskiy2021an_vit} and now to the physical sciences \citep{avsec2021effective, jumper2021highly}. In brief, an attention layer takes as input three matrices $K,Q,V$ in $\R^{L\times d}$, which should usually be thought of as length $L$ lists of $d$-dimensional vectors. The attention layer then outputs $Y=\sigma (Q K^\top )V\in \R^{L\times d}$ where $\sigma$ indicates a row-wise softmax operation. In the popular self-attention variant, $K,Q$ and $V$ are themselves learned functions (usually simple linear transformations) of a single input sequence $X=(x_1,\dots,x_L)^\top \in \R^{L\times d}$. Intuitively, one might imagine that this mechanism can model long range dependencies because the ``computational path'' between any row $x_i$ and any row $y_j$ is very short: more formally, this model avoids the ``vanishing gradients'' issue that plagues RNNs, in which the partial derivative $\frac{\partial y_L}{\partial x_1}$ decays exponentially in $L$. Unfortunately, this requires $O(L^2)$ time and space due to the need to construct the matrix $QK^\top \in \R^{L\times L}$, which places a limitation on how large $L$ can be.

The surge in popularity of attention has engendered a corresponding surge in interest on methods for increasing the context length $L$ while controlling the $O(L^2)$ computational cost. Broadly speaking, most approaches fall into two camps: those that attempt to ``linearize'' attention, and those that sparsify the attention matrix $QK^\top$. The first camp exploits the fact that if the softmax is removed, we can re-write attention as $Q (K^\top V)$, where now $K^\top V\in \R^{d\times d}$ and so the whole operation is only linear in $L$ rather than quadratic. Thus, if we could replace the softmax with a different operation that allowed some version of this rearrangement, it would be possible to speed up the attention layer. This idea is the underlying principle behind methods such as the Performer \citep{choromanski2020rethinking}, Linear Attention \citep{katharopoulos2020transformers}, Random Feature Attention \citep{peng2020random} or cosFormer \citep{qin2022cosformer}. In the other camp, one employs the simple expedient of simply not computing the entire matrix $QK^\top$, usually by employing some kind of sparsification strategy to only compute some of the elements of this matrix. For example, BigBird \citep{zaheer2020big}, GMAT \citep{gupta2020gmat}, Longformer \citep{beltagy2020longformer}, and Blockwise self-attention \citep{qiu2019blockwise} all choose particular sparsification patterns that are intuitively likely to capture "important" interactions, while the Reformer \citep{kitaev2020reformer} uses locality-sensitive hashing to dynamically adjust the sparsification pattern. Most recently, \citet{Hawthorne2022GeneralpurposeLA} compress the history in a small number of latents for long range autoregressive tasks. In essence, these approaches can be viewed as a trade-off between performance, and computation time (or memory).

Despite its current empirical success, the attention mechanism is not the only approach for modeling sequence data, nor even necessarily the most natural. For example, a classical \emph{recursive} or \emph{state space} layer operates on an input $(x_1,\dots,x_L)^\top\in \R^{d\times L}$ by defining a sequence of ``states'' $s_{t+1} = T(s_t, x_{t+1})$ and returning the output sequence $y_t = E(s_t)$ where $T$ and $E$ are learned ``transition'' and ``emission'' functions. Assuming $T$ and $E$ can be computed efficiently, this requires only $O(L)$ time. Note however, that the $O(L)$ theoretical complexity is complicated by the fact that these models appear to be inherently serial, while the $O(L^2)$ matrix operations in attention are much more easily parallelizable. State space models have additional attractive properties: they naturally capture our physical intuition that the universe is described by some appropriate state, and the output $y_t$ can be theoretically influenced by every single prior input, rather than requiring the model to commit to some context window of fixed length $L$. Through the years, many different possibilities for $T$ and $E$ have appeared, such as the simple RNN, which sets $T$ and $E$ to both be MLPs, or more complicated LSTM layer \citep{hochreiter1997long}. Nevertheless, the performance of state space models has in many cases been quickly surpassed by attention.

There have been several recent efforts to recapture the favorable properties of state space models while maintaining or improving upon the performance of attention models. For example, the Transformer-XL \citep{Dai2019transformerxl} modifies attention to incorporate state via a sliding-window style approach, while the Block-Recurrent Transformer \citep{hutchins2022block} directly parameterized the $T$ and $E$ functions using attention layers. Further, alternative models based on linear dynamical systems have shown great promise \citep{gu2022efficiently, dss_gupta2022diagonal,Gu2022OnTP}. Rather than attempting to ``fix'' attention, these are classical state space models where both $T$ and $E$ are linear. An important insight in these works is to recast the model as a convolution with a very large kernel, and then leverage the convolution theorem to compute the sequence $y_t$ in $O(L\log(L))$ time using Fast Fourier Transform \citep{Cooley1965AnAF}. While seemingly worse than $O(L)$, this operation is easily parallelizable, and so in practice is significantly faster. Moreover, in a situation in which the $\log(L)$ factor is onerous, one may still fall back to the serial $O(L)$ algorithm. Despite their apparent simplicity, these models have demonstrated remarkable success in tasks that require integrating information from distant parts of the input: the key innovation is that careful initialization and parameterization of the linear maps is required in order to train the models effectively. Our approach will build on these works.

\section{Method}\label{sec:ssm}


We start by reviewing the necessary background on state spaces required to fully describe our model (\S\ref{sec:state-spaces}). We then formally define our \lasso model in \S\ref{sec:lasso} and the \lasso-Transformer-Hybrid in \S\ref{sec:hybrid}.

\subsection{State Space Preliminaries}\label{sec:state-spaces}

While in this work we are interested in modeling sequences of vectors, let us first review how state space models define a sequence-to-sequence map for 1-D sequences.

\paragraph{State Spaces} A discretized state space, parameterized by a state matrix $A \in \R^{N \times N}$, vectors $B \in \R^{N \times 1}$, $C \in \R^{1 \times N}$\comment{, $D \in \R^{1 \times 1}$} and a sample time $\Delta \in \R_{> 0}$ defines a sequence-to-sequence map from input $(u_0,\ldots,u_{L-1}) = u \in \R^L$ to output $(y_0,\ldots,y_{L-1}) = y \in \R^L$ via the recurrence\footnote{Discretization used in Equation \ref{eqn:discrete} relies on zero order hold (ZOH) which assumes a constant value between sampled times. We also experimented with other discretization methods such as bilinear transform but did not see much difference in performance.},
\begin{equation}\label{eqn:discrete}
\begin{split}
&x_k = \bar{A}x_{k-1} + \bar{B}u_k\ \ \ ,\ \ \ y_k = \bar{C}x_k \comment{ + \bar{D}u_k}\\[5pt]
&\bar{A} = e^{A\Delta}\ \;,\ \bar{B} = (\bar{A} - I)A^{-1}B\ ,\ \;\bar{C} = C\comment{\,,\ \;\bar{D} = D}\ .
\end{split}
\end{equation}
Assuming $x_{-1} = 0$ for simplicity, the above recurrence can be explicitly unrolled as
\begin{equation}\label{eqn:unroll}
y_k \ = \ \sum_{j=0}^k \bar{C}\bar{A}^j\bar{B}\cdot u_{k-j}\ .
\end{equation} 
For convenience, the scalars $\bar{C}\bar{A}^k\bar{B}$ are gathered to define the SSM kernel $\bar{K} \in \R^L$ as
\begin{equation}\label{eqn:kernel}
\begin{split}
\bar{K} \ \ = \ \ (\bar{C}\bar{B}, \bar{C}\bar{A}\bar{B}, \ldots, \bar{C}\bar{A}^{L-1}\bar{B}) \ \ =\ \ (\ C e^{A\cdot k\Delta} (e^{A\Delta} - I)A^{-1}B\ )_{0 \leq k < L},
\end{split}
\end{equation}
where the last equality follows by substituting the values of $\bar{A}$, $\bar{B}$, $\bar{C}$ from Equation \ref{eqn:discrete}. Hence, 
\begin{equation}\label{eqn:unroll-kernel}
y_k \ = \ \sum_{j=0}^k \bar{K}_j\cdot u_{k-j}\ .
\end{equation}
where $\bar{K}_j$ denotes the value of the kernel at position $j$. 
Given an input sequence $u \in \R^L$, it is possible to compute the output $y \in \R^L$ sequentially via the recurrence in Equation \ref{eqn:discrete}. While this property is highly desirable for autoregressive decoding, a sequential computation is prohibitively slow to train with long inputs and, instead, Equation \ref{eqn:unroll-kernel} can be used to compute all elements of $y$ in parallel, provided we have already computed $\bar{K}$.

\paragraph{Computing $y$ from $u$ and $\bar{K}$ is easy.} Given an input sequence $u \in \R^L$ and the SSM kernel $\bar{K} \in \R^L$, naively using Equation \ref{eqn:unroll-kernel} for computing $y$ would require $O(L^2)$ multiplications. Fortunately, this can be done much more efficiently by observing that for the univariate polynomials 
$$\bar{K}(z) = \sum_{i=0}^{L-1} \bar{K}_i z^i \ \ \text{and}\ \ u(z) = \sum_{i=0}^{L-1} u_i z^i ,$$ 
$y_k$ is the coefficient of $z^k$ in the polynomial $\bar{K}(z) \cdot u(z)$, i.e. all $y_k$'s can be computed simultaneously by multiplying two degree $L-1$ polynomials. It is well-known that this can be done in $O(L\log(L))$ time via Fast Fourier Transform (FFT) \citep{cormen}. We denote this fast computation of Equation \ref{eqn:unroll-kernel} via the discrete convolution as
\begin{equation}\label{eqn:convolution}
y = \bar{K} *_\mathrm{c} u \ .
\end{equation}

\paragraph{Diagonal State Spaces} The challenging part is computing $\bar{K}$ itself as it involves computing $L$ distinct matrix powers (Equation \ref{eqn:kernel}). \cite{dss_gupta2022diagonal} observed that the state matrix $A$ can be assumed to be diagonal without loss in performance, thereby allowing a straighforward computation of $\bar{K}$. Their \dssexp model (\dss from hereon) assumes $A$ to be a diagonal matrix $\mathrm{diag}(\lambda_1, \ldots, \lambda_N)$ and assumes $B = (1)_{1 \leq i \leq N}$. \dss has parameters $\Lambda_\mathrm{re}, \Lambda_\mathrm{im} \in \R^N$, $C \in \C^N$ and $\Delta_{\log} \in \R$. The diagonal of $A$ (i.e. $(\lambda_1, \ldots, \lambda_N)$) is computed as $-\elemexp(\Lambda_\mathrm{re}) + i\cdot \Lambda_\mathrm{im}$ where $i = \sqrt{-1}$ and $\Delta$ is computed as $\mathrm{exp}(\Delta_{\log}) \in \R_{> 0}$. For this parameterization, the kernel (Equation \ref{eqn:kernel}) can be computed as a matrix-vector product
\begin{equation}\label{eqn:dss-kernel}
\bar{K}\ \ =\ \ (C * \left( (e^{\lambda_i\Delta} - 1) / \lambda_i \right)_{1\leq i \leq N})_{1\times N} \cdot \elemexp(P_{N\times L})
\end{equation}
where $P_{i,k} = \lambda_i k\Delta$ and * denotes elementwise multiplication.

This formally defines a \textit{linear} sequence-to-sequence map for 1-D sequences. In the case of sequences of $H$-dimensional vectors, state space models are applied individually on the $H$ features as follows. 

Each \dss layer receives a length-$L$ sequence $u$ of $H$-dimensional vectors as input, i.e., $u \in \R^{H\times L}$, and produces an output $y \in \R^{H\times L}$. The parameters of the layer are $\Lambda_\mathrm{re}, \Lambda_\mathrm{im} \in \R^N$, $\Delta_{\log} \in \R^{H}$ and $C \in \C^{H \times N}$. For each coordinate $h=1,\ldots,H$, a state space kernel $\bar{K}_h \in \R^L$ is computed for $\Lambda_\mathrm{re}, \Lambda_\mathrm{im}, (\Delta_{\log})_h, C_h$ via Equation \ref{eqn:dss-kernel}. The output $y_h \in \R^L$ for coordinate $h$ is computed from $u_h \in \R^L$ and $\bar{K}_h$ using Equation \ref{eqn:convolution}.

For batch size $B$, sequence length $L$ and hidden size $H$, the DSS layer requires $O(NHL)$ time to compute the kernels, and $O(BHL\log(L))$ time for the discrete convolution. In S4 and DSS, the authors suggest to use $N=64$ as default and for this choice of $N$, the time taken to compute the kernels becomes an important factor, specially for small batches.

\subsection{\lasso Layer}\label{sec:lasso}

Armed with the tools and techniques from the state space literature, we now turn to the main contribution of our work and describe our \lasso layer in detail (Figure \ref{fig:lasso}).

We build on the idea of the recently introduced Gated Attention Unit (GAU) \citep{flash_hua2022transformer} and replace the $\Omega(L^2)$ attention used in GAU by a further simplified \dss layer (\S\ref{sec:state-spaces}). Gating allows our model to be contextualized over a reduced dimensionality and the use of state spaces provides it with superior contextualizing abilities, while enjoying $O(L\log L)$ complexity out of the box.

Concretely, our \textit{Gated State Space} (``\lasso'') layer maps a sequence $X \in \mathbb{R}^{L \times E}$ to an output $O \in \mathbb{R}^{L \times E}$ as
\begin{align}
U &= \phi(XW_1) \ \in\ \R^{L \times H} \qquad \quad V = \phi(XW_2) \ \in \ \R^{L \times F} 
\label{eq:lasso1} \\
Y &= \dss(U) \ \in \ \R^{L \times H} \qquad \quad U_\mathrm{context} = YW_3 \ \in \ \R^{L \times F} 
\label{eq:lasso2} \\
O &= (U_\mathrm{context} * V)W_4 \ \in \ \R^{L \times E} 
\label{eq:lasso3}
\end{align}
where $*$ denotes elementwise multiplication and $\phi$ denotes $\mathrm{GELU}$ \citep{Hendrycks2016BridgingNA}. Note that, contrary to \cite{flash_hua2022transformer}, in our experiments, we did not observe much benefit from using $\mathrm{RELU}^2$ or Swish activations instead of $\mathrm{GELU}$.

In Equation \ref{eq:lasso2}, a contextualized representation of $U$ is formed via \dss in $O(NHL) + O(BHL\log L)$ time. In most of our experiments we used $H=E/4=256$, $N=512$ and $F=4096$. Similar to GAU, gating allows us to perform a weaker contextualization using the state space layer by reducing the dimension of the input to \dss, by $4\times$ in our case. This offers a much needed increase in throughput since we observed that FFT ops were the main bottleneck (measured on TPUs). As shown in Table \ref{tab:dss_lasso}, this provides more than $3\times$ speedups over the vanilla \dss block described in \S\ref{sec:state-spaces}.

\paragraph{Simpified DSS used in \lasso layer} In Equation \ref{eq:lasso2}, instead of directly using \dss as described in \S\ref{sec:state-spaces}, we made several key simplifications based on our exploratory results. 

As described in \S\ref{sec:state-spaces}, S4 and \dss train a separate $\Delta$ for each of the $H$ coordinates resulting in the creation of $H$ separate $P$ matrices in Equation \ref{eqn:dss-kernel}. In the \dss layer used in \lasso, we decided to eliminate the use of $\Delta$ by fixing it as $1$. This reduces the compute required for creating the kernels and makes the kernel computation simpler. 


Secondly, we found that randomly initializing the parameters corresponding to $\Lambda$ works just as well as the Skew-Hippo initialization suggested in \citep{dss_gupta2022diagonal}. In practice, we parametrize both real and imaginary parts of $\Lambda$ in log space (\S\ref{app:code})\comment{, similar to how $\Delta$ is initialized}. The effectiveness of random initialization is in contrast to the findings of \cite{gu2022efficiently} and \cite{dss_gupta2022diagonal} who reported their respective initializations of the state space parameters to be critical to the model performance. We do however note that the experiments in our setting of large-scale language modeling are conducted on orders of magnitude larger scale of compute than what is used in the tasks considered in these works. Moreover, the data modalities we consider in this work are different from the tasks considered in these works (e.g. Long Range Arena) which are specifically designed to stress test long-range reasoning.

\subsection{\lasso-Transformer-Hybrid}\label{sec:hybrid}
Conceptually, \lasso looks fairly different from the current workhorse of machine learning; the Transformer architecture. Given this, it is not immediately clear if one is decidedly better than the other or if they each provide some orthogonal benefits. If its the latter, one might wonder if there are synergies between these architectures which can be exploited to create a hybrid model which is stronger than either one of them individually. To that end, we also consider a conceptually simple hybrid between \lasso and Transformer where we sparingly interleave traditional Transformer blocks with \lasso layers. Despite its glaring simplicity, as shown in Table \ref{tab:other}, we observed that it shows consistent and significant improvements on all tasks.

\textbf{Chunking long inputs}\ \ \ In all our experiments we used sequence lengths large enough to be prohibitive for traditional Transformer layers. To get around this restriction at the Transformer layers used in our hybrid model, we chunk their inputs into non-overlapping chunks of length 512 and run the Transformer layer on each of them independently. While the \lasso layers are apt at modeling both short and longer range interactions, the interleaved Transformer layers can potentially allow for a richer modeling of short range interactions. 

\comment{
2. Maybe also mention length generalization in intro?
3. Maybe we can summarize the empirical results in a couple of lines in the intro - not super important.
4. maybe have an extended results table in Appendix containing other models from Hutchins et al to highlight we're competitive with models people are more aware of than Block recurrent transformers.
5. people tend to look at results table first - maybe bold the GSS or say ours?
6. use ICLR style file if you're aiming for it? Another option is TACL - submission is 1st of every month and 1 month anonymity (Aug 1st deadline for us). We dont have to agree to publish it even if it gets accepted but can give us valuable feedback and decent backup.
}

\section{Results}
We conduct experiments with \lasso on 4 different datasets, LM1B, PG19, ArXiv and Github, each of them varying qualitatively with another in terms of modality and average document length.

\textbf{LM1B} is a standard and reasonably big dataset (1B tokens) where each training example consists of a short sentence \citep{lm1b_DBLP:journals/corr/ChelbaMSGBK13}. This is different from rest of the datasets which consists of a much larger sequence of tokens per example. Although, our primary goal is to measure \lasso's ability to capture long range dependencies, we include results on LM1B benchmark and compare with vanilla Transformer baseline, which is hard to do for larger sequence lengths.

\textbf{PG19} dataset is constructed from extracting a large collection of full-length books from Project Gutenberg \citep{pg19_raecompressive2019}. All extracted books were written before 1919 and only contain tokens from the English language. Over the years PG-19 has become a standard benchmark for measuring progress on long range dependency modeling over text.

\textbf{ArXiv Math} dataset was recently collected by \cite{wu2022memorizing} and contains LaTeX source for articles focusing on Mathematics. Even though articles are typically shorter than full-length books, as in PG19, since LaTeX source can have many special characters, typical sub-piece vocabularies are not very effective and tends to produce examples of similar size, as measured by number of tokens. It is possible to train a custom sub-piece vocabulary for this dataset, but we stick with the vocabulary used by both \cite{wu2022memorizing} and \cite{hutchins2022block} for fair comparison.

\textbf{Github} was also first collected and used by \cite{wu2022memorizing}. It is a corpus of raw source code collected from several Github repositories with open-source licences. The dataset contains code from several programming languages, including C, C++, Java, Python, Go and Typescript. In this case individual files can be small but code repositories are typically is organized so that code can be reused across file boundaries. Thus, for every repository, all the files were concatenated to produce a single document. 

We do token level modeling on all the datasets and report resulting perplexity numbers on a heldout set of examples. Perplexity numbers are obtained using teacher forcing (or parallel mode) where the correct output from the heldout set is used for decoding the next token at each position. For a fair comparison with the baselines we considered, we keep the vocabularies consistent as used by the baselines models. Specifically, we used custom trained 30k sized sentence-piece vocab for LM1B, T5 vocab with 32k tokens for PG19 \citep{raffel2019exploring} and Meena vocab with 32k tokens \citep{meena_adiwardana2020humanlike} for both ArXiv and Github datasets. 

\subsection{Comparing DSS and \lasso}
\label{subsec:lasso_dss}
\begin{table*}[t]
    \centering
   \begin{tabular}{lllc|cccc}
    \toprule
            \addlinespace[0.1cm]
        & & & & \multicolumn{4}{c}{Eval Sequence Length} \\
        \addlinespace[0.1cm]
        \hline
         \addlinespace[0.1cm]
        Dataset & Model & Params & Throughput (steps/sec) & \multicolumn{4}{c}{Perplexity} \\
        \addlinespace[0.1cm]
        \hline
        \addlinespace[0.1cm]
         & & & & 512 & 4k & 16k & 65k   \\
        \midrule
     \multirow{ 3}{*}{LM1B}   & Transformer & 182M & 6.6 & 13.51 \\
                                &  DSS  & 188M & 1.8 & 13.59 &&& \\
                                & \texttt{\lasso}  & 190M & 5.6 & 13.26 &&& \\
                                \hline
                               \addlinespace[0.1cm]
    \multirow{ 2}{*}{PG-19} &DSS & 209M & 1.8 & 14.51 & 13.53 & 692.1 & OOM \\
                               &\texttt{\lasso}  & 192M & 5.3 & 14.01 & 12.84 & 12.94 & 12.47 \\
                               \hline
                               \addlinespace[0.1cm]
    \multirow{ 2}{*}{Arxiv}     &DSS & 209M & 1.8 & 3.65 & 3.13 & 284.6 & OOM \\
                                &\texttt{\lasso}  & 192M & 5.3 & 3.57 & 3.08 & 3.08 & 2.75 \\
                                \hline
                                \addlinespace[0.1cm]
     \multirow{ 2}{*}{Github}    &DSS & 209M & 1.8 & 3.65 & 3.21 & 242.7 & OOM \\
                                &\texttt{\lasso}  & 192M & 5.3 & 2.68 & 2.35 & 2.31 & 2.12 \\
         \bottomrule
    \end{tabular}
        \caption{Comparison of DSS and \lasso models in fixed-param setting. We consistently find that \lasso outperforms DSS (with hyperparameters taken from \cite{dss_gupta2022diagonal}) while being 2-3$\times$ faster on all tasks. Training sequence length is 4k, except for LM1B for which it is 512.}
            \label{tab:dss_lasso}
\end{table*}
As shown in Table \ref{tab:dss_lasso}, in our first set of results, we compare DSS and \lasso models both in terms of perplexity and throughput, as measure by steps per second on all 4 datasets. Except for LM1B, all models were trained with 4k sequence length at training time. However, at evaluation time, we present results for on a large range of sequence lengths. For LM1B, we train and evaluate with sequence length of 512, since the dataset only contains short sentences, most examples comfortably fully fitting in. Despite being a (slightly) smaller model, \lasso performs quite favorably as measured by perplexity while being 2-3$\times$ faster depending on the dataset. 

Furthermore, we observe significant generalization over changes in sequence length on all the datasets. Not only does the performance not degrade when increasing sequence length, it actually improves quite a bit! This suggests that the model is effective at utilizing the extra context even though the model was not trained with that amount of context. Note that we used default hyperparameters suggested by \cite{dss_gupta2022diagonal} for initializing state space variables for DSS. But, for \lasso, since we made several structural changes, we retuned the hyperparameters related to initialization on PG19 alone and used them for all the datasets. Since we had access to evaluation metrics for all sequence lengths, length generalization factored in for our choice of hyperparameters, which is not true for DSS. It is completely possible we see similar length generalization even with DSS if initialization hyperparameters were retuned.

\textbf{Training details}\ \ All the models were trained with $2^{19}$ tokens per batch and 125k total training steps. We make sure to change the batch size as a function of the sequence length so that number of tokens in the batch remains the same. For example, for LM1B we set batch size to 1024 and sequence length to 512 but for rest of the datasets we use batch size of 128 and sequence length of 4k. For datasets with longer documents, we also considered increasing the sequence length even further. Intuitively, training on longer sequences would help the model learn longer range dependencies better. On the flip side, it makes the optimization a bit more challenging due to large number of correlated tokens per batch and would even likely result in overfitting. Since \lasso is able to generalize beyond the length it was trained on in most cases, we found sequence length of 4k to be a reasonable middle ground for this set of experiments.

Similar to language modeling experiments in \citep{gu2022efficiently}, every block of DSS baseline consists of DSS layer followed by GLU \citep{glu_dauphin2016language} and a Feedforward layer similar to the one used in Transformer block with \gelu activation \citep{Hendrycks2016BridgingNA}. DSS baseline consists 12 layers and an embedding dimension of 1024. For \lasso, we increased the number of layers to 16 to match the parameter count.


\subsection{Comparison with other baselines}
\begin{table*}[t]
    \centering
   \begin{tabular}{lll@{\hskip 20pt}l|cccc}
    \toprule
            \addlinespace[0.1cm]
        & & & & \multicolumn{4}{c}{Eval Sequence Length} \\
        \addlinespace[0.1cm]
        \hline
         \addlinespace[0.1cm]
        Dataset & Model & Params & {\hskip -15pt} TPUv4 hours & \multicolumn{4}{c}{Perplexity} \\
        \addlinespace[0.1cm]
        \hline
        \addlinespace[0.1cm]
         & & & & 512 & 4k & 16k & 65k   \\
        \midrule
    \multirow{ 2}{*}{PG-19} & \texttt{Rec:fixed:skip} & 196M & 0.8k & & 11.55 &  & \\
                            & \texttt{Feedback:lstm:single} & 196M+ & 0.8k+ & & 11.31 &  & \\
                            & \texttt{Feedback:fixed:skip} & 196M+ & 0.8k+ & & 11.24 &  & \\
                            & \textbf{\lasso}\hfill(\textbf{this work})   & 192M & 0.5k & 14.01 & 12.84 & 12.94 & 12.47 \\
                            & \textbf{\lasso}-L   & 352M & 0.8k & 12.48 & 11.33 & 11.16 & 11.12 \\
                            & \textbf{\lasso}-Hybrid-L   & 373M & 0.8k & 11.45 & \textbf{10.52} & 10.44 & 10.1 \\
                               \hline
                               \addlinespace[0.1cm]
    \multirow{ 2}{*}{Arxiv}  & \texttt{Rec:fixed:skip} & 196M & 0.8k & & 2.36 &  & \\
                            & \texttt{Feedback:lstm:single} & 196M+ & 0.8k+ & & \textbf{2.33} &  & \\
                            & \texttt{Feedback:fixed:skip} & 196M+ & 0.8k+ & & 2.36 &  & \\
                            & \lasso  & 192M & 0.5k & 3.57 & 3.08 & 3.08 & 2.75 \\
                            & \lasso-L  & 352M & 0.8k & 3.29 & 2.71 & 2.72 & 2.51 \\
                            & \lasso-Hybrid-L  & 373M & 0.8k & 2.94 & 2.51 & 18.27 & 125.2 \\
                                \hline
                                \addlinespace[0.1cm]
     \multirow{ 2}{*}{Github} & \texttt{Rec:fixed:skip} & 196M & 3k & & 2.04 &  & \\
                            & \texttt{Feedback:lstm:single} & 196M+ & 3k+ & & 2.07 &  & \\
                            & \texttt{Feedback:fixed:skip} & 196M+ & 3k+ & & 2.16 &  & \\
                            & \lasso  & 192M & 0.5k & 2.68 & 2.35 & 2.31 & 2.12 \\
                            & \lasso-L  & 352M & 1.8k & 2.31 & 1.99 & 2.20 & 2.28 \\
                            & \lasso-Hybrid-L  & 373M & 1.8k & 2.34 & \textbf{1.88} & 1.74 & 2.09 \\
         \bottomrule
    \end{tabular}
        \setlength{\belowcaptionskip}{-10pt}
        \caption{Comparing \lasso variants with best-performing models from \cite{hutchins2022block} in both fixed-param and fixed-compute settings. 
        While, in the fixed-param setting, \lasso currently lags behind performance of best Block Recurrent Transformer baseline, it is fairly competitive in the fixed-compute setting (as measured by total TPUv4 hours taken to complete training).
        For the larger model \lasso-L used for fixed compute comparison, we simply double the layers from 16 to 32 keeping everything else fixed. In addition, due to the inherent recurrent view of state space model family, decoding at inference time is much faster than Transformer based baselines. For block-recurrent baselines, adding feedback increases both parameter count and training time, we stick with conservative estimates derived from \cite{hutchins2022block} for both, which we denote by `+'. \cite{hutchins2022block} report $\log_2$ of perplexity which we convert to raw perplexity in this table. Param count for all the models include embedding layers as well.}
            \label{tab:other}
\end{table*}
In this section, we turn our attention towards apples-to-apples comparison of \lasso versus well-tuned baselines on these datasets. For a complete picture of the cost of training these models, we report both the number of parameters and time spent training as measured by TPUv4 hours. For baselines, we selected the best performing models reported in \citep{hutchins2022block} for every dataset and compare with \lasso model both in fixed-param and fixed-compute settings. 

\textbf{Training Details} \ \ Similar to Section \ref{subsec:lasso_dss}, unless otherwise mentioned, all the non-baseline models were trained using 64 TPUv4 cores for 125k steps. We use batch size of 128 and 4k sequence length at training time, with a total of $2^{19}$ tokens per batch. We increase the batch size to 256 for the Github dataset (with a token count of $2^{20}$ per batch) since we observed a lot of noise in our metrics. This is consistent with observations made by \cite{hutchins2022block}. 

We used Adam optimizer \citep{kingma2014adam} and tuned the base learning rate over a grid of $\in$ $[0.0064, 0.0032, 0.0016, 0.0008]$. We also employ linear warmup for 1k steps and cosine decay until 1e-6. We also observed better performance by using a higher than typical weight decay rate of 0.1, other than that we did not employ any additional regularization techniques, including dropout. Note that similar to \citep{gu2022efficiently,dss_gupta2022diagonal}, we used a constant learning rate of 1e-3 and set weight decay rate to 0.0 for state space parameters part of \lasso Layer . In addition, we clip the gradient to norm 1.0 before passing to the optimizer. Both of these helped with certain instabilities we observed in our preliminary experiments. 

\textbf{\lasso models} \ \ \lasso consists of 16 layers and an embedding dimension of 1024. We also consider a larger variant with 32 layers as denoted by \lasso-L. For \lasso-Hybrid model, we used vanilla Transformer blocks at every 4th layer starting with the 2nd layer. Since \lasso layers are inherently position aware, using them for the 1st layer eschews any need of explicit position embeddings typically used with otherwise position invariant Transformer blocks. Thus, barring position aware nature of \lasso layers, we don't use any kind of explicit position embedding or bias in our models. For the Transformer blocks used in hybrid models, we use multi-head self-attention with 8 heads, each with size 128. 

\textbf{Baselines} \ \ We considered 3 high-performing baselines and numbers reported in \citep{hutchins2022block}. Block Recurrent Transformer leverages recurrence over blocks of Transformer layer to model long range dependencies. \cite{hutchins2022block} performed a comprehensive exploration of open design space of incorporating recurrence across blocks. Somewhat surprisingly, \texttt{Rec:fixed:skip}, which accumulates recurrent state vector as an exponential moving average over time performs better than more complicated gating designs. Another variation which performed well with Block Recurrence is the idea of adding a feedback mechanism over blocks similar to Feedback Transformer \citep{feedback_fan2020addressing}. Note that feedback mechanism makes training more expensive due to additional cross-attention modules and corresponding paramaters.

\textbf{Fixed-param comparison} \ \ As shown in Table \ref{tab:other}, we see that \lasso variants come very close but not quite beat the strongest block recurrent baseline in the fixed-param setting. In this case, we are comparing \lasso model which has roughly 192M parameters (including embeddings) with the baselines all of which have around 196M parameters. Even though the parameter count is fixed, we see that \lasso runs faster than block recurrent baselines, likely due to the fact that all the layers can be completely parallelized unlike the the recurrence in the baselines which run in a sequential fashion.

\textbf{Fixed-compute comparison} \ \ Since the \lasso model runs faster than the baselines, we also train with versions larger than \lasso such that the training time (as measured by total TPUv4 hours) matches the time taken by the baselines. We simply double the number of layers from 16 to 32 to construct \lasso-L. As expected, adding more parameters improves perplexity numbers on the eval set of all the datasets. Moreover, we find that the \lasso-Hybrid versions of the model outperform the best baseline model on PG-19 and Github datasets. We do see significant improvements for Arxiv dataset as well but unfortunately not enough to be stronger than the baseline. We think this may be resolved by the use of a vocabulary more suited to Arxiv symbols. On the flip side, we can no longer do a fair comparison with token level perplexity if we change the vocabulary, so we stick with the vocabularies used by the baselines for this study.

\textbf{Length generalization} \ \ We train all variants of \lasso with sequence length of 4k but evaluate on 4 different lengths $l \in [512, 4k, 16k, 65k]$. On PG-19, we see significant length generalization across the board. Not only the performance doesn't degrade as the sequence length is increased but it gets significantly better for all model variants. On Arxiv and Github, the situation is a little more complicated. For smaller models, we still see length generalization but it tends to degrade when either the model is made bigger or the dataset has a lot of noise (as indicated by variation in perplexity metric over time). How to robustly achieve length generalization is an interesting research question on it is own and we believe one can design interventions which can lead to further improvements, which we don't explore here. 

Note that block recurrent baselines, with or without feedback mechanism, process documents in a sequential fashion such that recurrent state from previous segment from the document is passed to the next segment, with backward pass truncated to the current segment. This means that, even though the segment sequence length is set to 4k, block recurrent models have (arguably weak) access to almost the entire past. Thus, perplexity comparison at sequence length 4k is slightly unfair towards \lasso models since they do \emph{not} employ such state caching mechanisms.

\comment{
\section{Conclusion}
\ag{maybe omit this section for arxiv version?}
We introduce GSS, a general purpose sequence model which leverages gated units and trains significantly faster as shown on several language modeling benchmarks. Further comparison with well tuned Transformer baselines suggests that GSS is fairly competitive in fixed-compute setting. We further show that a hybrid model constructed by interleaving GSS and Transformer improves performance even further.
}

\begin{ack_harsh}
We are grateful to the developers of Jax and Flax libraries. In addition, we would like to thank DeLesley Hutchins and Imanol Schlag for answering our questions, sharing the datasets and helping us in general with details of Block Recurrent Transformer paper. Finally, we are grateful to Ethan Dyer for discussions on long context modeling,  Walid Krichene and John Anderson for providing feedback on an earlier draft of this paper. 
\end{ack_harsh}



{
\small
\bibliographystyle{plainnat}
\bibliography{ref}
}

\newpage
\appendix

\section{Supplemental Material}\label{sec:supplemental}

\subsection{Fast convolution via FFT}

For $u, v \in \C^{1\times L}$ the Circular Convolution Theorem states that, 
\begin{align*}
\mathrm{invFFT}_L(\mathrm{FFT}_L(u) * \mathrm{FFT}_L(v)) = 
v \cdot {\begin{bmatrix} 
u_{0} & u_1 & \cdots & u_{L-1} \\
u_{L-1} & u_0 & \ddots & \vdots \\
\vdots & \ddots & \ddots & u_{1} \\
u_{1} & \cdots & u_{L-1} & u_{0} \\
\end{bmatrix}} = v \cdot \mathrm{circulant}(u)\ .
\end{align*}
where $*$ denotes elementwise multiplication. As $\mathrm{FFT}, \mathrm{invFFT}$ can be done in $O(L\log L)$ time this provides a fast algorithm for circulant matrix-vector product. In practice, linear systems can often be expressed as a circulant matrix-vector product and is also true in the case of Equation \ref{eqn:unroll-kernel} which can be equivalently expressed as 
\begin{align*}
[y_{0}\  \ldots\ y_{L-1}\ |\ 0\ \ldots\ 0]_{1\times 2L} = 
[\bar{K}\ | \ 0\ \ldots\ 0]_{1\times 2L} \cdot \mathrm{circulant}([u_{0}\  \ldots\ u_{L-1}\ |\ 0\ \ldots\ 0])_{2L\times 2L}\ .
\end{align*}

\subsection{Implementation of \lasso}\label{app:code}

\begin{figure*}[h]
\centering
\begin{minipage}{.99\textwidth}
\begin{minted}
[
% frame=lines,
framesep=2mm,
baselinestretch=1.1,
fontsize=\footnotesize,
% linenos
]
{python}
    
def simplified_dss_kernel(H, L, N=512):
    # Lambda_re, Lambda_im: [N]
    # C_re, C_im: [H N]
    Lambda = -Lambda_re.exp() + 1j*Lambda_im.exp()  # [N]
    C = C_re + 1j*C_im                              # [H N]
    S = (Lambda * arange(L).view(1,L)).exp()        # [N L]
    C = C * (Lambda.exp() - 1) / Lambda             # [H N]
    return einsum('hn,nl->hl', C, S).real           # [H L]

def dss(u, H, L):
    u = norm(u)
    # compute H state space kernels
    K = simplified_dss_kernel(H, L)             
    K_f = rfft(K, pad_to=2*L)      
    u_f = rfft(u, pad_to=2*L)      
    y = irfft(K_f * u_f)[...,:L] 
    # param D: [H,1]
    return y + D * u

def gss(x, F=4096, L=4096, E=1024, H=256):
    shortcut, x = x, norm(x)
    v = dense(x, F, activation='gelu')
    u = dense(x, H, activation='gelu') 
    y = dss(u, H, L)
    uc = dense(y, F)
    o = dense(uc * v, E)
    return o + shortcut
\end{minted}
\end{minipage}
\caption{Pseudocode of \lasso (\S\ref{sec:lasso}).}\label{fig:code}
\end{figure*}

\end{document}